\pdfoutput=1

\documentclass[11pt]{article}

\usepackage{ACL2023}

\usepackage{times}
\usepackage{latexsym}
\usepackage{multirow}
\usepackage{tabularx}
\usepackage{todonotes}
\usepackage{graphicx}
\usepackage{tipa}
\usepackage[T1]{fontenc}

\usepackage[utf8]{inputenc}

\usepackage{microtype}

\usepackage{inconsolata}

\title{Parallel Corpus for Indigenous Language Translation: Spanish-Mazatec and Spanish-Mixtec}

\author{
\normalsize Atnafu Lambebo Tonja$^{1}$, Christian Maldonado-Sifuentes$^{2}$, David Alejandro Mendoza Castillo$^{2}$, \\
\textbf{\normalsize Olga Kolesnikova$^{1}$, Noé Castro-Sánchez$^{3}$,  Grigori Sidorov$^{1}$, Alexander Gelbukh$^{1}$}  \\
\footnotesize
$^1$Instituto Politécnico Nacional (IPN), Centro de Investigación en Computación (CIC), Mexico,\\
\footnotesize
$^2$Transdisciplinary Research for Augmented Innovation -  Laboratory (TRAI-L), Mexico, \\
\footnotesize
$^3$Departamento de Ciencias Computacionales Tecnológico Nacional de México, Mexico \\
}
\begin{document}
\maketitle
\begin{abstract}
In this paper, we present a parallel Spanish-Mazatec and Spanish-Mixtec corpus for machine translation (MT) tasks, where Mazatec and Mixtec are two indigenous Mexican languages. We evaluated the usability of the collected corpus using three different approaches: transformer, transfer learning, and fine-tuning pre-trained multilingual MT models. Fine-tuning the Facebook M2M100-48 model outperformed the other approaches, with BLEU scores of 12.09 and 22.25 for Mazatec-Spanish and Spanish-Mazatec translations, respectively, and 16.75 and 22.15 for Mixtec-Spanish and Spanish-Mixtec translations, respectively. The findings show that the dataset size (9,799 sentences in Mazatec and 13,235 sentences in Mixtec) affects translation performance and that indigenous languages work better when used as target languages. The findings emphasize the importance of creating parallel corpora for indigenous languages and fine-tuning models for low-resource translation tasks. Future research will investigate zero-shot and few-shot learning approaches to further improve translation performance in low-resource settings. The dataset and scripts are available at \url{https://github.com/atnafuatx/Machine-Translation-Resources}. 

\end{abstract}

\section{Introduction}
Natural Language Processing (NLP), a sub-field of Artificial Intelligence (AI), has been attracting a lot of attention in terms of research and development as a result of the surge in the number of applications it has in a variety of different industries \cite{kalyanathaya2019advances}. Machine Translation (MT), Sentiment or Opinion Analysis, POS Tagging, Question Classification (QC) and Answering (QA), Chunking, Named Entity Recognition (NER), Emotion Detection, and Semantic Role Labeling are currently highly researched areas in various high-resource languages \cite{tonja2023natural}.

The domain of machine translation (MT) is advancing at a rapid pace due to the growing prevalence of computational tasks and the expanding global reach of the Internet, which caters to diverse, multilingual communities \cite{kenny2018machine}. MT systems have demonstrated remarkable translation outcomes for language pairs that possess abundant resources, such as English-Spanish, English-French, English-Russian, and English-Portuguese. However, in scenarios with limited or no resources, MT systems encounter difficulties due to the primary obstacle of inadequate training data for certain languages \cite{mager2018challenges,tonja2021parallel,tonja2022improving, tonja2023low}.

Low-resource languages have been suffering from a lack of new language technology designs. When the resources are limited and only a small amount of unlabeled data is available, it is very hard to reach a true breakthrough in creating powerful novel methods for language applications \cite{tonja2022improving}, the problem becomes worse if there is no parallel dataset for certain languages. 

Mexico is a multicultural and multilingual country with 68 officially recognized indigenous languages, 238 variants, and Spanish, a widely used language spoken by 90 percent of the population \cite{mager2021findings}. Few language technologies have been developed for indigenous languages spoken in Northern and Southern America; moreover, many indigenous languages spoken in the Americas face a risk of extinction \cite{mager2018challenges}.

Indigenous language speakers often experience feelings of shame or reluctance to use their native languages, primarily due to limited opportunities for application in the presence of pervasive, dominant majority languages \cite{hornberger2008, skutnabb2000}. This phenomenon can be attributed to social and cultural pressures that prioritize the use of majority languages over minority languages, thereby marginalizing indigenous linguistic communities and undermining the value of their linguistic heritage \cite{hinton2011}. 

In this paper, we introduce the first parallel corpus for machine translation tasks for two indigenous languages that are spoken in Mexico and benchmark experimental results. The contributions of our work are the following:
\begin{itemize}
    \item We introduce the first parallel corpus for machine translation for Mazatec and Mixtec languages.
    \item We evaluate the performance of the collected corpus and present benchmark results by using transformers, transfer learning, and fine-tuning approaches. 
    \item We open-source the parallel corpus and the scripts used in this paper.
\end{itemize}
The rest of the paper is organized as follows: Section \ref{related} describes previous research related to this study, Section \ref{lang} describes the properties of Mazatec and Mixtec languages, Section \ref{dataset} describes the statistics of the collected dataset, Section \ref{baseline} describes models used for baseline experiments and their results, and Section \ref{Conc} describes the conclusion of the paper.

\section{Related works} \label{related}
Due to an increase in the enormous amount of data for different languages, machine translation is currently one of the most researched areas in NLP and has shown promising results in high-resource languages \cite{tonja2022improving}. There are different MT approaches that have been used by different researchers, neural machine translation (NMT)is one of the current state-of-the-art approaches trained on huge datasets containing sentences in a source language and their equivalent target language translations \cite{belay2022effect}. Basically, NMT takes advantage of huge translation memories with hundreds of thousands or even millions of translation units \cite{forcada2017making}. However, NMT for low-resource languages still under-performs due to the scarcity of parallel datasets \cite{tonja2022improving,tonja2023low}.

Many researchers explored different approaches to solving low-resource machine translation problems. \citet{zoph2016transfer} proposed a transfer learning method to improve the MT performance of low-resource languages. The authors first train a high-resource language pair (the parent model), then transfer some of the learned parameters to the low-resource pair (the child model) to initialize and constrain training. The data augmentation approach proposed by \citet{fadaee2017data}, targets low-frequency words by generating new sentence pairs containing rare words in new, synthetically created contexts. \citet{pourdamghani2019neighbors} proposed using high-resource language resources to improve MT performance for low-resource languages without requiring any parallel data. Copying monolingual data of the target language is proposed by \citet{currey2017copied} to improve the performance of low-resource MT.
\citet{tonja2023low} proposed the use of source-side monolingual data as another way of improving low-resource MT performance. Transfer learning method, where one first trains a "parent" model for a high-resource language pair and then continues training on a low-resource pair only by replacing the training corpus was proposed by \citet{kocmi2018trivial}. Mixing low-resource language resources during training, as proposed by \citet{tonja2022improving} showed an improvement in MT performance for low-resource languages.

There have been promising research works done for indigenous languages; \citet{feldman2020neural} presented an NMT model and a dataset for the Bribri Chibchan language for Bribri-Spanish translation. \citet{kann2022americasnli} compiled AmericasNLI, a natural language inference dataset covering 10 indigenous languages of the Americas. They conducted experiments with pre-trained models, exploring zero-shot learning in combination with model adaptation. \citet{oncevay2021peru} proposed the first multilingual translation models for four languages spoken in Peru: Aymara, Ashaninka, Quechua, and Shipibo-Konibo, providing both many-to-Spanish and Spanish-to-many models, outperformed pairwise baselines. \citet{zheng2021low} presented a low-resource MT system that improves translation accuracy using cross-lingual language model pre-training. The authors used an mBART implementation of fairseq to pre-train on a large set of monolingual data from a diverse set of high-resource languages before fine-tuning on 10 low-resource indigenous American languages: Aymara, Bribri, Asháninka, Guaraní, Wixarika, Náhuatl, Hñähñu, Quechua, Shipibo-Konibo, and Rarámuri. On average, their proposed system achieved BLEU scores that were 1.64 higher and chrF scores that were 0.0749 higher than the baseline. \citet{nagoudi2021indt5} introduced IndT5, the first Transformer language model for 10 Indigenous American languages: Aymara, Bribri, Asháninka, Guaraní, Wixarika, Náhuatl, Hñähñu, Quechua, Shipibo-Konibo, and Rarámuri. To train IndT5, they built IndCorpus--a new dataset for ten indigenous languages and Spanish. 

\section{Languages} \label{lang}
\subsection{Mazatec}  \label{mazatec}
The Mazatec language comprises a collection of closely related indigenous languages spoken primarily in the Northern region of Oaxaca, with smaller populations in the adjacent states of Puebla and Veracruz in Mexico. Approximately 200,000 individuals speak Mazatec; however, this number may fluctuate depending on which particular dialects or linguistic variations are taken into account \cite{Leonard2019mazatec}.

Mazatec belongs to the Oto-Manguean language family, a large family of indigenous Mesoamerican languages which also includes Mixtec, Zapotec, Otomi, among others \cite{vielma2017panorama}. Linguistic characteristics of Mazatec include tonal distinctions \cite{garellek2011acoustic}, complex consonant clusters, and a rich morphology \cite{leonard2012almaz}. The Mazatec languages are known for their agglutinative structure, where words are formed by combining multiple morphemes, each with a distinct meaning \cite{vielma2017panorama}.
\subsubsection{Writing system} \label{writing}
\textbf{Vowels -}
Mixtec has five basic vowels, similar to those in Spanish:

\begin{itemize}
\item a (as in "car"), \item e (as in "bet"),
\item i (as in "bit"), \item o (as in "bore"),
\item u (as in "boot").
\end{itemize}

These vowels can also appear nasalized, indicated by a tilde ($\tilde{a}$, $\tilde{e}$, $\tilde{i}$, $\tilde{o}$, $\tilde{u}$), and long, indicated by a colon ($a: , e: , i: , o: , u: $). Tones can be associated with vowels, too.

\textbf{Consonants -}
The Mazatec consonant inventory includes the following sounds:
\begin{itemize}
\item Stops: p, t, k, b, d, g,
\item Affricates: ts, t\textesh, dz, d\textyogh,
\item Fricatives: s, \textesh, h, z, \textyogh,
\item Nasals: m, n, ng,  
\item Approximants: w, j (pronounced as "y" in "yes"),
\item Lateral approximant: l,
\item Rhotics: r.
\end{itemize}

\textbf{Numerals/Numbers -}
Mazatec uses a vesimal numeral system (base-20). Here are the numbers 1 to 10 in Mazatec:
(1) - ki\underline{a}, (2) - chj\textbari{}, (3) - tsi, (4)- sti, (5) - nk\underline{a}, (6)  - tsj\textbari{}, (7) - kja, (8)  - chj\textbari{}n, (9)  - ts\underline{i}, (10) - st\underline{i}.

\textbf{Word order -}
Typically, Mazatec exhibits a VSO (Verb-Subject-Object) word order; however, alternative structures such as SVO can also occur depending on the sentence, the focus of the statement, and the context.

\textbf{Example sentence:}

Kitsaara kji xi makjíñeni kua apana (I gave a pill for the headache to my father) - VSO order

\subsection{Mixtec}

The Mixtec language comprises a group of closely related indigenous languages predominantly spoken in the region known as La Mixteca, which spans the states of Oaxaca, Puebla, and Guerrero in Southern Mexico. Estimates indicate that there are approximately 500,000 speakers of Mixtec; however, this number may fluctuate depending on the specific dialects or language varieties considered \cite{josserand1983mixtec}.

As Mazatec, Mixtec is a member of the Oto-Manguean language family \cite{rensch1977otomanguean,pike1961language,hollenbach2000mixtec} possessing the characteristic mentioned in Section \ref{mazatec}. It also shares the phonemic system with Mixtec (see vowels and consonants inventory in Section \ref{writing}) as well as the word order features and the base-20 number system.  
Here are numbers from 1 to 10 in Mixtec:
(1)- in, (2) - ña'a, (3) - ta'a , (4)  - na'a, (5)  - ma'a, (6)  - chiko, (7)  - chikue, (8)  - chikuiin, (9)  - chikunña'a, (10)  - ndo'o.  


\textbf{And here are a couple of examples sentences:}
\begin{itemize}
\item Ka'nu ña'a nuu ntaa (Sitting on the plain) - VSO order
\item Ña'a nuu ntaa ka'nu (On the plain, sitting) - SVO order
\end{itemize}

Note that the Mixtec language has many dialects, so the phonetic inventory, numerals, word order, and example sentences provided here may vary across different Mixtec-speaking communities. The examples given here are intended to provide a general overview of the language's features

\section{Parallel Dataset} \label{dataset}
Data is one of the crucial building blocks of any NLP application \cite{belay2022effect,tonja2023natural}, and a parallel corpus is essential to the success of any machine translation task. For Mazatec and Mixtec, we were unable to find publicly available datasets for the MT task. We collected datasets for these two indigenous Mexican languages from two main domains: \textit{religious} and \textit{constitution}. We also collected additional resources for the Mixtec language from different \textit{textbooks} which have a similar translation to Spanish. Table \ref{sentsize} shows the statistics of the collected parallel corpus for Mazatec and Mixtec. 

\textbf{Text Alignment -} We took a base directory path where text files were stored as input. Then we read and merged the content of all text files in the directory, and obtained a list of lists containing the content of each file. We proceeded to iterate through each file in the directory and read their contents line by line. Each line was normalized using the Unicode Normalization Form KC (NFKC) before being appended to the resulting list. We added a function that takes a language code \texttt{lang} as input, which determines the filename of the text file to be read from a predefined folder. The function read the file line by line, normalized each line using NFKC, and concatenated the lines into a single string. The result was returned as an array.

With another function, we added the two lists as input: one containing the content of the files to be aligned, and the other containing the filenames for the output files. We then iterated through the content list and aligned the text by iterating through the chapters and paragraphs of each translation. The aligned text was written to the corresponding output file as tab-separated values (TSV). Then we defined the root path where the input files were located, initialized the name and content arrays, and called the function that populated the content array with the pre-processed text. Finally, the function that writes the file was called to align and write the output files.

\begin{table*}[ht]
\small
\begin{tabular}{l|ll|l|ll|l}
\hline
\multirow{2}{*}{Source} & \multicolumn{3}{l|}{\textbf{Mazatec (maq) - Spanish (spa)}} & \multicolumn{3}{l}{\textbf{Mixtec (xtn) - Spanish (spa)}} \\ \cline{2-5} 
                        & \multicolumn{1}{l|}{\#sentences}   & \#tokens (maq)& \#tokens (spa)  & \multicolumn{1}{l|}{\#sentences}   & \#tokens (xtn) & \#tokens (spa)\\ \hline
Religion      & \multicolumn{1}{l|}{8,203} & 269,753& 187,773& \multicolumn{1}{l|}{8,208} & 278,874 &183,050\\ \hline
Constitution & \multicolumn{1}{l|}{1,596} &138,504  & 68,392& \multicolumn{1}{l|}{1,185} &104,497 &68,393 \\ \hline
Others        & \multicolumn{1}{l|}{-} &-  &- & \multicolumn{1}{l|}{3,842} &71,628 & 70,080\\ \hline
Total        & \multicolumn{1}{l|}{9,799} &408,257 & 256,165 & \multicolumn{1}{l|}{13,235} & 454,999 &321,523\\ \hline
\end{tabular}
\caption{Parallel dataset distribution of Mazatec-Spanish and Mixtec-Spanish \label{sentsize}}
\end{table*}

\textbf{Pre-processing -} After aligning the texts of two indigenous languages with their equivalent translations in Spanish, we pre-processed the corpus before splitting it for our experiments. The pre-processing steps included removing the numbers and special character symbols such as ;,",?, etc. For the baseline experiment, we split the pre-processed corpus into training, development, and test sets in the ratio of 70:10:20, respectively. Table \ref{datasplit} shows the split of the dataset used for our experiments.
\begin{table}[ht]
\begin{tabular}{l|lll}
\hline
\multirow{2}{*}{\textbf{Language pairs}} & \multicolumn{3}{l}{\textbf{Number of Sentences}}                                       \\ \cline{2-4} 
                                         & \multicolumn{1}{l|}{\textbf{Train}} & \multicolumn{1}{l|}{\textbf{Dev}} & \textbf{Test} \\ \hline
Mazatec - Spanish & \multicolumn{1}{l|}{7,056} & \multicolumn{1}{l|}{784} & 1,959  \\ \hline
Mixtec - Spanish & \multicolumn{1}{l|}{9,529} & \multicolumn{1}{l|}{1,059} &2,647  \\ \hline
\end{tabular}
\caption{Dataset split used in baseline experiments \label{datasplit}}
\end{table}

\section{Baseline Experiment and Discussion} \label{baseline}
In this section, we discuss the models used for the baseline experiment, the hyper-parameter used, the benchmark results, and the discussion. We used three approaches to evaluate the usability of the collected corpus. These are :-
\begin{itemize}
    \item \textbf{Transformer} - is a type of neural network architecture first introduced in the paper \textit{Attention Is All You Need} \cite{vaswani2017attention}. 
    The key innovation of the Transformer architecture is the attention mechanism, which allows the network to selectively focus on different parts of the input sequence when making predictions. This is in contrast to traditional recurrent neural networks (RNNs), which process input sequentially and are prone to the vanishing gradient problem.

   In the transformer architecture, the input sequence is processed in parallel by multiple layers of self-attention and feed-forward neural networks. Each layer can be thought of as a "block" that takes the output of the previous layer as input and applies its own set of transformations to it. The self-attention mechanism allows the network to weigh the importance of each element in the input sequence when making predictions, while the feed-forward networks help to capture non-linear relationships between the elements.

    Currently, transformers are state-of-the-art approaches and are widely used in NLP tasks such as MT, text summarization, sentiment analysis, etc. We used the base transformer configuration as described in \cite{vaswani2017attention} work.
    
    \item \textbf{Transfer learning}- refers to the process of leveraging pre-trained language models to improve the performance of downstream NLP tasks. Specifically, transfer learning involves using a pre-trained model to initialize the parameters of an MT system and then fine-tuning the system on a smaller dataset specific to the target language pair or domain.
    
    Transfer learning can be especially useful in MT because training a high-quality MT system from scratch requires a large amount of data and computational resources, which may not be available for all language pairs or domains. By leveraging pre-trained models, transfer learning allows MT systems to achieve high performance with fewer data and fewer resources.
    For our baseline experiments, we used English-Spanish as parent model with two (\textbf{opus-mt-es-en}\footnote{https://huggingface.co/Helsinki-NLP/opus-mt-es-en} and \textbf{opus-mt-tc-big-en-es }\footnote{https://huggingface.co/Helsinki-NLP/opus-mt-tc-big-en-es}) pre-trained models available from Hugging Face\footnote{https://huggingface.co/} trained for English-Spanish on the OPUS dataset \cite{tiedemann-thottingal-2020-opus} by \textit{Helsinki-NLP group}. 
    
    \item \textbf{Fine tuning} - is the process of taking a pre-trained MT model and adapting it to a specific translation task, such as translating between a particular language pair or in a specific domain. The process of fine-tuning involves taking the pre-trained model, which has already learned representations of words and phrases from a large corpus of text, and training it on a smaller dataset of specific task examples. This involves updating the parameters of the pre-trained model to better capture the patterns and structures present in the target translation task.
   
   Fine-tuning can be useful in MT because it allows the pre-trained model to quickly adapt to a new task without having to train a new model from scratch. This is especially beneficial when working with limited data or when there is a need to quickly adapt to changing translation requirements. 
   We used two commonly known pre-trained multilingual MT models: 
   \begin{itemize}
       \item \textbf{M2M100-48} -  is a multilingual encoder-decoder (seq-to-seq) model trained for many-to-many multilingual translation \cite{fan2020englishcentric}. We used a model with 48M parameters due to computing resource limitations.
       \item \textbf{mBART50} - is a multilingual sequence-to-sequence model pre-trained using the \textit{Multilingual Denoising pre-training} objective \cite{tang2020multilingual}. 
   \end{itemize} 
\end{itemize}
\textbf{Hyper-parameters - } For the transformer approach we tokenized the source and target parallel sentences into	subword tokens using Byte Pair Encoding (BPE) \cite{gage1994new}. The BPE representation was chosen in order to remove vocabulary overlap during dataset combinations. For other approaches we applied the tokenizer of each model, Table \ref{tab:param} shows hyper-parameters used in our baseline experiments. 
\begin{table*}[h!]
\centering
\begin{tabular}{l|l|l}
\hline
\textbf{Approaches}           & \textbf{Models}                   & \textbf{Parameters} \\ \hline
Transformer &
  transformer &
  \begin{tabular}[c]{@{}l@{}}- enc\_layers: 6\\ - dec\_layers: 6\\ - heads: 8\\ - hidden\_size: 512\\ - optimizer: adam\\ - warmup\_steps: 4000\\ - training\_steps: 30000\\ - learning \_rate: 5e-2\end{tabular} \\ \hline
\multirow{2}{*}{Transfer learning} &
  opus-mt-es-en &
  \multirow{4}{*}{\begin{tabular}[c]{@{}l@{}}- max\_seq\_length: 128\\  - num\_train\_epochs: 3\\ - per\_device\_batch\_size:  4\\ - num\_beams: 5\end{tabular}} \\ \cline{2-2}
                             & opus-mt-tc-big-en-es &                     \\ \cline{1-2}
\multirow{2}{*}{Fine-tuning} & mBART50                 &                     \\ \cline{2-2}
                             & M2M100-48                   &                     \\ \hline
\end{tabular}
\caption{Hyper-parameters used for baseline experiments \label{tab:param}}
\end{table*}

\subsection{Results}
\begin{figure*}[h!]
    \centering
    \includegraphics[width=\linewidth]{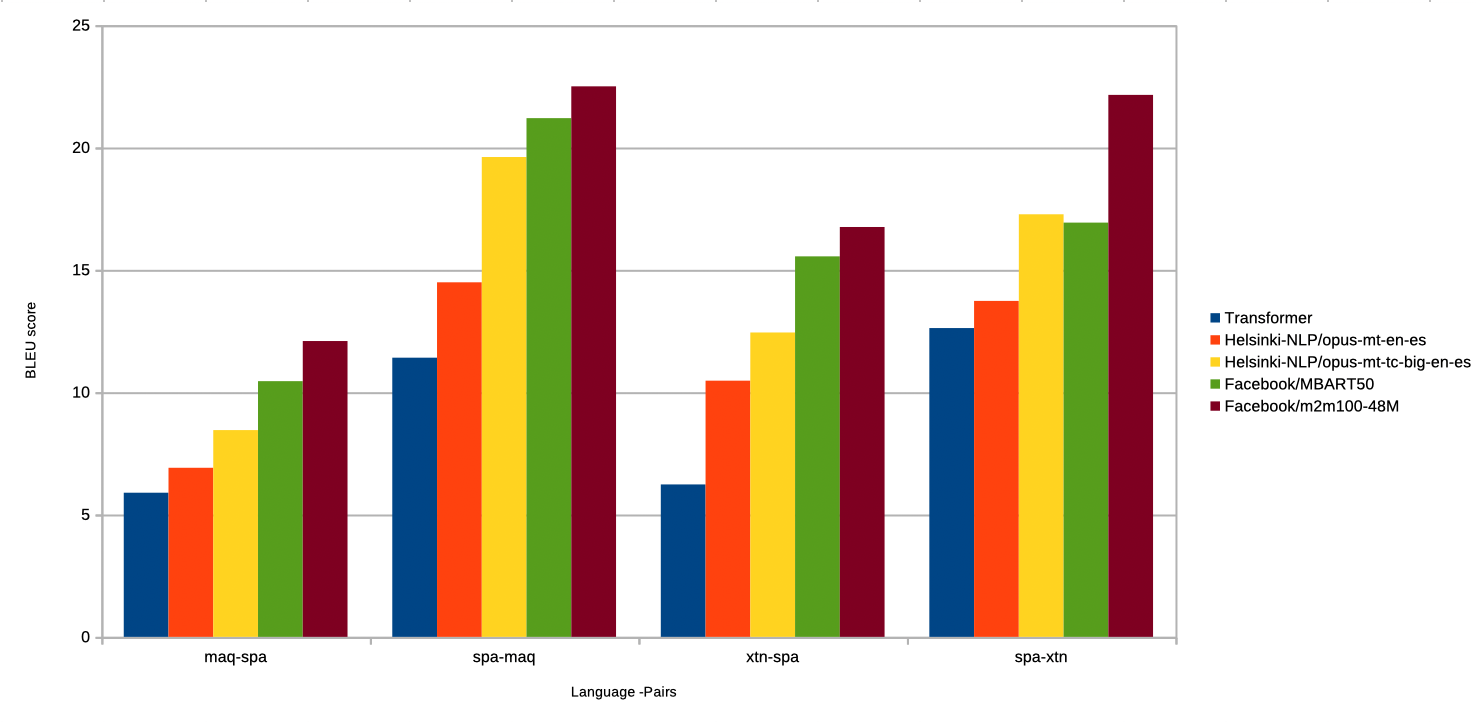}
    \caption{Benchmark results of selected approaches}
    \label{fig:my_label}
\end{figure*}

Table \ref{results} and Figure \ref{fig:my_label} shows the benchmark experimental results for bi-directional neural machine translation for Mazatec(maq) - Spanish(spa) and Mixtec(xtn) - Spanish(spa). 
\begin{table*}[h!]
\centering
\begin{tabular}{l|ll|ll}
\hline
\multirow{2}{*}{\textbf{Models}} & \multicolumn{2}{l|}{\textbf{xx-spa BLEU score}}          & \multicolumn{2}{l}{\textbf{spa-xx BLEU score}}          \\ \cline{2-5} 
                                 & \multicolumn{1}{l|}{\textbf{maq-spa}} & \textbf{xtn-spa} & \multicolumn{1}{l|}{\textbf{spa-maq}} & \textbf{spa-xtn} \\ \hline
\textbf{M1} & \multicolumn{1}{l|}{5.89} &6.23  & \multicolumn{1}{l|}{11.41} & 12.62\\ \hline
\textbf{M2}        & \multicolumn{1}{l|}{6.91} &10.47  & \multicolumn{1}{l|}{14.49} &13.73  \\ 
\textbf{M3}        & \multicolumn{1}{l|}{8.45} &12.44  & \multicolumn{1}{l|}{19.61} &17.27  \\ \hline 
\textbf{M4}     & \multicolumn{1}{l|}{10.45} &15.66  & \multicolumn{1}{l|}{21.2} &  16.93 \\ 
\textbf{M5}  & \multicolumn{1}{l|}{\textbf{12.09}} & \textbf{16.75} & \multicolumn{1}{l|}{\textbf{22.5}} & \textbf{22.15} \\ \hline

\end{tabular}
\caption{Benchmark experimental result for bi-directional Mazatec(maq)-Spanish(spa) and Mixtec(xtn)-Spanish(spa) neural machine  translation, M1, M2, M3, M4, and M5 represents transformer, opus-mt-es-en, opus-mt-tc-big-en-es,mBART50, and M2M100-48 models respectively.
  \label{results}  }
\end{table*}
In our baseline experiments, we observed that employing a transformer model for low-resource languages shows sub-optimal results compared to transfer learning and fine-tuning methodologies. As demonstrated in Table \ref{results} and Figure \ref{fig:my_label}, the performance of the \textbf{transformer} was inferior to alternative approaches utilized in the study. This finding substantiates the hypothesis that the efficacy of transformer models is heavily reliant on the availability of extensive parallel corpora for machine translation tasks. Upon further examination of language pair performance, we discovered that utilizing indigenous languages as the target language surpasses the performance achieved when using Spanish as the target language. This observation indicates that translating from Spanish to indigenous languages is a less complex task for the model as opposed to translating indigenous languages to Spanish.

\textbf{Transfer learning} approach showed more promising results for the indigenous low-resource languages than the transformer approach. Out of the two models used in the transfer learning experiment, the model with \textit{transformer-big} configuration outperformed the model with \textit{transformer-base} configuration. This shows that the transfer learning approach depends on the size of the model parameter. Similarly, when using the transfer learning approach for indigenous low-resource languages by utilizing models trained on high-resource languages, better results were obtained when Spanish was used as the source language than when Spanish was used as the target language.

\textbf{Fine-tuning} approach outperformed the rest of the approaches used in our baseline experiment in both translation directions. This shows that using a multilingual pre-trained translation model for fine-tuning low-resource languages outperforms other models. From the two multilingual models used in the experiment, the \textbf{M2M100-48} model outperformed the \textbf{mBART50} multilingual model. The M2M100-48 model showed 4.7 and 5.5 BLEU scores on average for Mazatec (maq)-Spanish (spa) and Spanish (spa)-Mazatec (maq) translation. For Mixtec (xtn)-Spanish (spa) and Mixtec (xtn)-Spanish (spa), the M2M100-48 model showed a 10.2 and 7.5 BLEU score improvement on average when compared to the other models used in the experiments. When comparing the results of the two languages in all the approaches used, Mixtec (xtn)-Spanish (spa) translation showed better performance than Mazatec (maq)-Spanish (spa) translation when using Spanish as the target language, This shows that the availability of the parallel corpora for the language pairs has a high impact on the performance of the translation models. The overall results show that using multilingual MT models for fine-tuning in our selected indigenous low-resource languages gives promising results.

\subsection{Discussion}
In our analysis, we conducted an error analysis to identify the strengths and weaknesses of the three approaches: transformer, transfer learning, and fine-tuning. We found that the transformer approach, which relies on large parallel corpora, yielded sub-optimal results for low-resource languages. It struggled to capture the linguistic patterns and structures specific to indigenous languages. This limitation indicates that the transformer model's performance is highly dependent on the availability of extensive parallel corpora for effective machine translation.

On the other hand, the transfer learning approach showed more promising results for low-resource indigenous languages. We observed that models pre-trained on high-resource languages, such as Spanish, and fine-tuned on the indigenous languages improved translation quality. However, even with transfer learning, the performance was not satisfactory, and there were errors that persisted across all three approaches.

The general error that all three approaches failed to address adequately was the translation of domain-specific and culturally specific terms in Mazatec and Mixtec. These languages have unique vocabulary and cultural nuances that require a deeper understanding and context to ensure accurate translation. The limited availability of domain-specific parallel corpora for these languages hampered the models' ability to capture and translate such terms effectively.

\section{Conclusion} \label{Conc}
In this paper, we presented a parallel corpus for two indigenous Mexican languages (Mazatec (maq) and Mixtec (xtn)) for machine translation tasks and evaluate the usability of the collected corpus using three different approaches. From the approaches, fine-tuning multilingual pre-trained MT models outperformed the rest of the experiments; Facebook's M2M100-48 outperformed all other models with BLEU scores of 12.09 and 22.25 for maq-spa and spa-maq, respectively, and 16.75 and 22.15 for xtn-spa and spa-xtn, respectively. We noticed from the experimental results that the dataset size has less impact when using indigenous languages as a target than the source. This observation highlights the potential benefits of focusing on developing and fine-tuning models specifically designed for translation tasks involving low-resource languages. Moreover, it underscores the value of creating and employing parallel corpora tailored to indigenous languages, as these resources can significantly improve machine translation performance, particularly when used in conjunction with advanced multilingual pre-trained models.

Our BLEU results for Mizatec and Miztec to Spanish translation were very low on the best configuration to have any usability in real-life applications, but the translation in the opposite direction demonstrated BLEU scores above 22 facilitating uses, for example in government apps to present hints to Mixtec and Mazatec native speakers who have a low level of Spanish comprehension, in the government web pages. This could significantly improve the usefulness of the native language of the speakers, thus promoting communication of the language and its preservation.

In future research, we plan to investigate the efficacy of advanced techniques, including zero-shot and few-shot learning, for low-resource languages in the context of limited parallel datasets. These methodologies hold promise for effectively leveraging sparse data available in low-resource settings, as they capitalize on pre-existing knowledge from related tasks or languages without requiring extensive fine-tuning or additional annotated data. By exploring these approaches, we aim to uncover potential benefits and improvements in the machine translation performance of low-resource languages, thus contributing to developing more robust and accurate translation systems for underrepresented linguistic communities.
\section*{Acknowledgements}

The work was done with partial support from the Mexican Government through the grant
A1S-47854 of CONACYT, Mexico, grants 20220852, 20220859, and 20221627 of the Secretaría de Investigación y Posgrado of the Instituto Politécnico Nacional, Mexico. The authors thank the CONACYT for the computing resources brought to them through the Plataforma de Aprendizaje Profundo para Tecnologías del Lenguaje of the Laboratorio de Supercómputo of the INAOE, Mexico, and acknowledge the support of Microsoft through the Microsoft Latin America PhD Award.
\bibliography{anthology,custom}
\bibliographystyle{acl_natbib}

\appendix



\end{document}